\documentclass{article}
\usepackage{spconf,graphicx}
\usepackage{amsmath,amsfonts}
\usepackage{algorithmic}
\usepackage{algorithm}
\usepackage{array}
\usepackage{textcomp}
\usepackage{stfloats}
\usepackage{url}
\usepackage{verbatim}
\usepackage{graphicx}
\usepackage{epstopdf}
\usepackage{cite}
\usepackage{adjustbox}
\usepackage{booktabs}
\usepackage{hyperref}
\hypersetup{hidelinks}
\usepackage{multirow}
\usepackage{booktabs}
\graphicspath {{ figures/ }}

\title{EFFECTIVE IMAGE TAMPERING LOCALIZATION VIA ENHANCED TRANSFORMER AND CO-ATTENTION FUSION}
\name{Kun Guo$^{1,2}$, Haochen Zhu$^{1,2}$, Gang Cao$^{1,2}$$^{\ast}$ \thanks{*Corresponding author}}
\address {\normalsize{$^{1}$State Key Laboratory of Media Convergence and Communication, Communication University of China, Beijing 100024, China}\\
\normalsize{$^{2}$School of Computer and Cyber Sciences, Communication University of China, Beijing 100024, China}}

\begin{document}
%
\maketitle
\begin{abstract}
Powerful manipulation techniques have made digital image forgeries be easily created and widespread without leaving visual anomalies. The blind localization of tampered regions becomes quite significant for image forensics. In this paper, we propose an effective image tampering localization network (EITLNet) based on a two-branch enhanced transformer encoder with attention-based feature fusion. Specifically, a feature enhancement module is designed to enhance the feature representation ability of the transformer encoder. The features extracted from RGB and noise streams are fused effectively by the coordinate attention-based fusion module at multiple scales. Extensive experimental results verify that the proposed scheme achieves the state-of-the-art generalization ability and robustness in various benchmark datasets. Code will be public at \href{https://github.com/multimediaFor/EITLNet}{https://github.com/multimediaFor/EITLNet}.
\end{abstract}
\section{Introduction}
\label{sec:intro}

Nowadays, digital image forgery has been drawing ever-increasing attention in our lives. As convenient editing tools are available, digital images can be tampered easily. Such tampered images are visually indistinguishable from original images and may lead to malicious use \cite{wu2019mantra, dong2022mvss}. Therefore, to fight against image forgery, it is crucial to accurately locate the tampered regions. Among many manipulation types, splicing, copy-move, and object removal are the three common semantic tampering techniques that have been studied most \cite{htsencar2022multimedia}. Splicing  means to insert a region copied from a different image while copy-move  refers to replicating an object from the same image. Removal, also known as inpainting, involves removing selected object region by extending the background. Such content-changed operations may cause serious misunderstandings. Generally, to better create realistic forgeries, some non-content-changed manipulations are also applied, such as brightness adjustment, Gaussian blur and JPEG compression. Since those manipulations do not affect the semantic information expressed in the image scene, our study focuses on three main types of image manipulation techniques, splicing, copy-move, and removal.

Many deep learning-based approaches have been proposed for image forgery localization with different backbones. Noiseprint \cite{cozzolino2019noiseprint} leverages Siamese network to capture the imaging model-related artifacts. Zhou \textit{et al}. \cite{zhuo2022self} present a two-stream network that consists of RGB and noise residual branches based on Faster R-CNN. Full convolution networks are used to capture specific or generic forensic clues \cite{wu2019mantra,wu2022robust,cozzolino2019noiseprint}. Furthermore, different attention modules are adopted to focus on tampered regions in a target image \cite{dong2022mvss,zhuo2022self,liu2022pscc}. In MVSS-Net \cite{dong2022mvss}, dual attention strategy \cite{fu2019dual} is applied to feature fusion at the late stage. A forgery attention module is integrated to a coarse-to-fine network \cite{zhuo2022self}. Liu \textit{et al}.\cite{liu2022pscc}  exploits spatial-channel correlation attention mechanism in a progressive network. 

Note that the prior methods often simply concatenate the RGB and noise view branches at an early or late stage for fusion, and ignore the interaction between two modalities. Additionally, attention modules are predominantly deployed in the decoder \cite{liu2022pscc} or at a single scale \cite{dong2022mvss}, so they are not integrated well into the localization network. To attenuate such deficiency of existing works, we propose a new effective image tampering localization network (EITLNet) with high generalization performance and robustness. It leverages both high and low levels of operation trace features by an enhanced two-branch transformer network with co-attention feature fusion. A feature enhancement (FE) module is used to strengthen the feature representation ability of encoder. The coordinate attention-based fusion (CAF) module is designed to integrate the RGB and noise feature streams effectively across multiple scales. It promotes the interlacing and calibration of complementary information between the two branch sub-networks. Extensive evaluation results verify the superior accuracy of EITLNet on benchmark datasets and their online social network processing versions. 

The remainder of the paper is organized as follows. In Section II, we present our proposed scheme in detail. Followed by extensive experimental results and discussions in Section III. We draw the conclusions in Section IV.

\begin{figure*}[h]
    \centering
    \includegraphics[width=\textwidth]{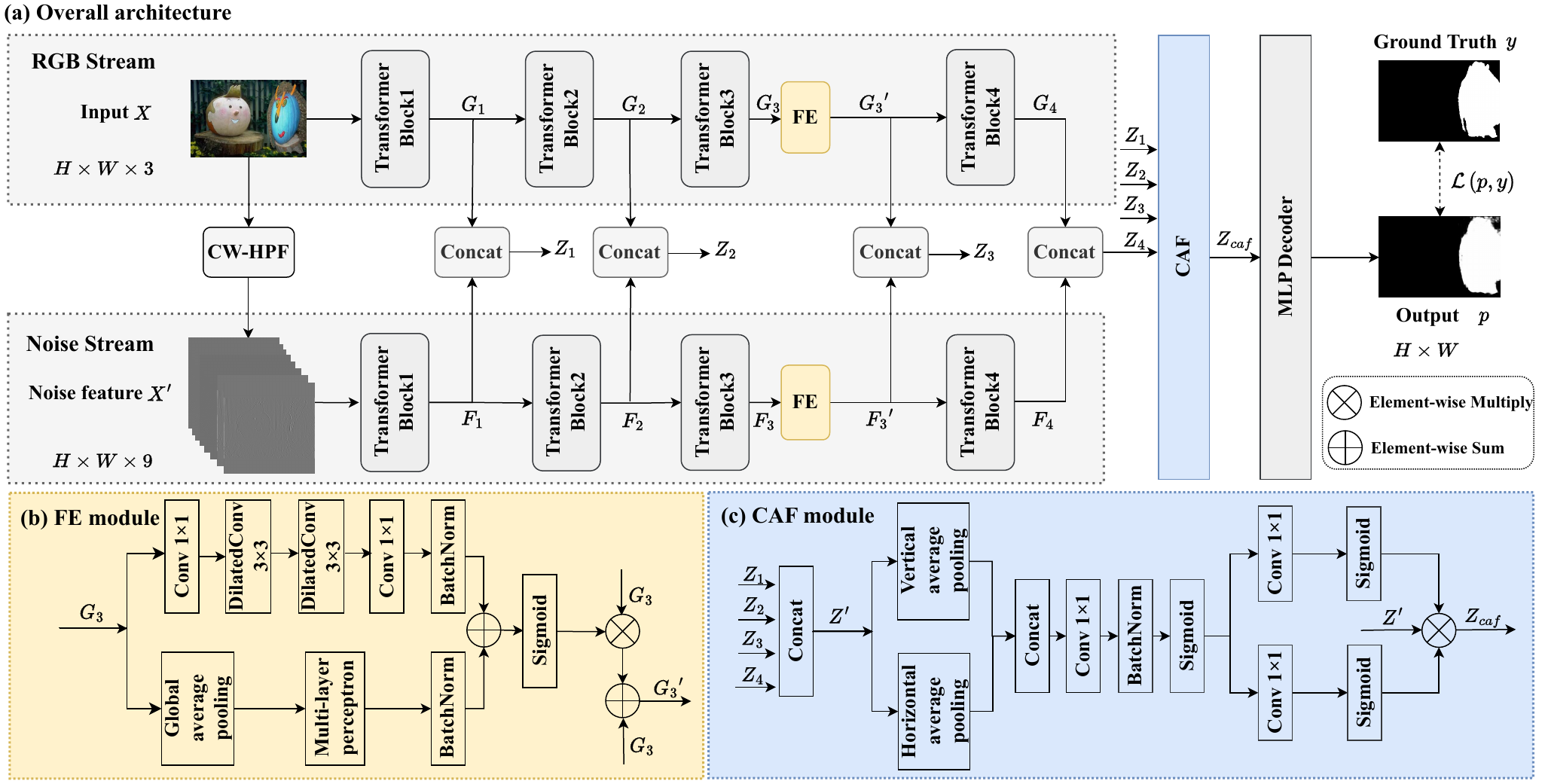}
    \caption{Architecture of the proposed EITLNet.}
    \label{Fig.1.}
\end{figure*}

\section{PROPOSED SCHEME}

\subsection{Overview}

Overview of the proposed tampering localization scheme is illustrated in Fig. 1. The Segformer \cite{xie2021segformer} network is used as backbone. To capture forensic clues, the RGB and noise features are extracted by two branches, each with four transformer blocks and a FE module. The transformer blocks are based on the Mix Transformer encoder Mit-B2. Features at each scale are finally fused with a CAF module for enhancing the interaction between RGB and noise models. Meanwhile, we adopt a multilayer perceptron (MLP) as decoder, which consists of an upsampling layer and a linear layer. Additionally, channel-wise high pass filters (CW-HPF) \cite{zhuo2022self} is used to extract noise with enlarged inter-channel inconsistency.

Specifically, we input an image $X$  of size $H\times W \times3$, as shown in Fig. 1., $X$ passes through the CW-HPF block and outputs a noise feature map $X'$ of size $H \times W \times 9$. $X$ and $X'$ are input to their corresponding feature-extraction branches. Then the feature maps at four levels are denoted as ${Z_n},\left( { n = 1,2,3,4} \right) $ with resolutions of 1/4, 1/8, 1/16, 1/32 of the original image respectively. We concatenate these four different scale feature maps correspondingly with a CAF module. Finally, the fused features are put into MLP decoder and predict a mask $P \in {\mathbb{R}^{H \times W}}$. The overview of our network is conceptually expressed as
\begin{equation}
P \leftarrow {\text{MLP}}\left( {{\text{CAF}}\left( {{Z_1},{Z_2},{Z_3},{Z_4}} \right)} \right)
\end{equation}

\subsection{Feature enhanced module}
To improve the feature representation ability of transformer encoders, an enhancement module is applied to the feature maps extracted by the middle transformer block. Such deployment is aimed at leveraging the high frequency features extracted from proper level of layers.
As pointed out  in \cite{yang2021multi}, the features extracted from shallow transformer layers are so noisy that the performance may degraded. The features from deep layers have less high frequency information.

Detailed architecture of the feature enhancement module is in Fig. 1 (b). Given the feature map ${G_3} \in {\mathbb{R}^{{H_3} \times {W_3} \times {C_3}}}$ yielded by the middle transformer block3 in RGB stream, the dilated convolutions $\operatorname{conv} _{3 \times 3}^{\operatorname{dila} }$ \cite{yu2015multi} are used to enlarge the receptive fields\cite{park2018bam}. Two $1 \times 1$ convolutions ${{\mathop{\rm conv}\nolimits} _{1 \times 1}}$ are used to adjust the number of feature map channels. In detail, the spatial attention map ${M_s}\left( {{G_3}} \right)$ is generated as
\begin{equation}
 \begin{aligned}
M_{s}\left(G_{3}\right)=&\mathrm{BN}\left(\mathrm{conv}_{\mathrm{l\times l}}\right.
\\&\left.\left(\mathrm{conv}_{3\times3}^{\mathrm{dik}}\left(\mathrm{conv}_{3\times3}^{\mathrm{dik}}\left(\mathrm{conv}_{\mathrm{l\times l}}\left(G_{3}\right)\right)\right)\right)\right)
\end{aligned}   
\end{equation}
where BN denotes batch normalization.
Furthermore, the inter-channel relationship is leveraged to form the channel attention. To aggregate the feature map in each channel, global average pooling (GAP) is applied to  ${G_3}$ and produce a ${\mathbb{R}^{1 \times 1 \times {C_3}}}$ channel feature map. Then MLP is adopted followed a batch normalization layer. That is,
\begin{equation}
    {M_c}\left( {{G_3}} \right) = \operatorname{BN} \left( {\operatorname{MLP} \left( {\operatorname{GAP} \left( {{G_3}} \right)} \right)} \right)
\end{equation}
Then we combine the ${M_s}\left( {{G_3}} \right)$  and  ${M_c}\left( {{G_3}} \right)$  by element-wise summation and take a sigmoid function to obtain the final attention map $M\left( {{G_3}} \right)$ as 
\begin{equation}
    M\left( {{G_3}} \right) = \operatorname{sigmoid} \left( {{M_s}\left( {{G_3}} \right) + {M_c}\left( {{G_3}} \right)} \right)
\end{equation}
Finally, the enhanced feature map ${G_3}^\prime$  is acquired as
\begin{equation}
    {G_3}^\prime  = {G_3}{\text{ + }}{G_3} \otimes M\left( {{G_3}} \right)
\end{equation}
where $\otimes$ denotes element-wise multiplication. The corresponding enhanced feature map ${F_3}^\prime$ is also yielded by applying the same module to the feature map ${F_3}$ of noise branch.

\subsection{Coordinate attention-based fusion module}
As illustrated in Fig. 1 (c), the features of two branches are first concatenated to be ${Z'}{\text{ =  [}}{Z_1},{Z_2},{Z_3},{Z_4}]$. Then average pooling in horizontal and vertical directions are used to aggregate features respectively, as 
\begin{equation}
    {Z'}\left( h \right) = \mathrm{HAP}({Z'})
\end{equation}
\begin{equation}
    {Z'}\left( w \right) = \mathrm{VAP}({Z'})  
\end{equation}
\noindent
where HAP is the horizontal average pooling and VAP is the vertical average pooling \cite{hou2021coordinate}. The spatial information is encoded horizontally and vertically as $T$ in Eq. (8). 
\begin{equation}
    T{\text{ = }}\operatorname{sigmoid} \left( {\operatorname{BN} \left( {{{\operatorname{conv} }_{1 \times 1}}\left( {\left[ {{Z^\prime}\left( w \right),{Z^\prime}\left( h \right)} \right]} \right)} \right)} \right)
\end{equation}
\noindent
where $\left[  \cdots  \right]$ denotes the concatenate operation. Then we split $T$ along the spatial dimension to two separate tensors, ${T^h}$ and ${T^w}$.  Such are expanded and used as attention maps. The attention map $M{_h}$ is yielded as 
\begin{equation}
    M{_h} = \operatorname{sigmoid} \left( {{{\operatorname{conv} }_{1 \times 1}}\left( {{T^h}} \right)} \right)
\end{equation}
\noindent
The $M{_w}$ is yielded likewise. Finally, the fused feature ${Z_{caf}}$ from CAF module is acquired as
\begin{equation}
{Z_{caf}}  = {Z'} \otimes M{_h} \otimes M{_w}
\end{equation}

\subsection{Loss function}
First, we employ the dice loss \cite{wei2021learn} used commonly in semantic segmentation networks is defined as
\begin{equation}
    \mathcal{L}_{Dice}(p,y) = 1 - \frac{{2\sum\nolimits {{p_i} \cdot {y_i}} }}{{\sum\nolimits {{p_i}^2}  + \sum\nolimits {{y_i}^2} }}
\end{equation}
where ${p_i}$ and ${y_i}$ are the prediction labels and ground truth for each pixel of the sample image respectively.
Then, to address the issue of imbalanced positive and negative sample distribution. The focal loss \cite{lin2017focal} is defined as
\begin{equation}
\begin{aligned}
       \mathcal{L}_{Focal}(p,y) =  &- {\sum {\alpha (1 - {p_i})} ^\gamma }{y_i}\log ({p_i}) \\ &- {\sum {(1 - \alpha ){p_i}} ^\gamma }(1 - {y_i})\log (1 - {p_i})
\end{aligned}
\end{equation}
where the hyperparameters $\gamma$ and $\alpha$ are the weights, for making the model focus on learning from more difficult samples. 

Overall, the final loss function is defined as   
\begin{equation}
    \mathcal{L}(p,y) = \mathcal{L}_{Focal}(p,y) + \mathcal{L}_{Dice}(p,y)
\end{equation}

\section{Experiment}
\subsection{Experimental settings}
\textbf{Training Dataset.} Following the prior work \cite{zhueffective}, a total of 60971 synthetic tampered images are used to train the proposed EITLNet. They are collected from a public synthetic image datasets \cite{bappy2017exploiting} and CASIAv2 \cite{dong2013casia}. The synthetic forged images are created by splicing the pristine images with the object regions from MS-COCO \cite{lin2014microsoft}. 

\noindent
\textbf{Testing Dataset.} Six commonly-used standard datasets are adopted for testing, including Coverage \cite{wen2016coverage}, CASIA-v1 \cite{dong2013casia}, Columbia \cite{hsu2006detecting}, NIST16 \cite{guan2019mfc}, DSO \cite{de2013exposing} and IMD \cite{novozamsky2020imd2020}. 

\noindent
\textbf{Implementation Details.} All images are resized to $512 \times 512$ pixels, and then enforced by general data augmentations including flip, scaling, Gaussian blur and JPEG compression. The network is initialized by the weights pretrained on ImageNet and trained for 100 epochs with the batch size 8 and a learning rate with cosine decay from 5e-3 to 5e-4. An AdamW optimizer is used with a momentum of 0.9.  The related parameters are set empirically as $\alpha=0.5$ and $\gamma=2$.
\begin{figure}[!t]
    \centering
    \includegraphics[width=\linewidth]{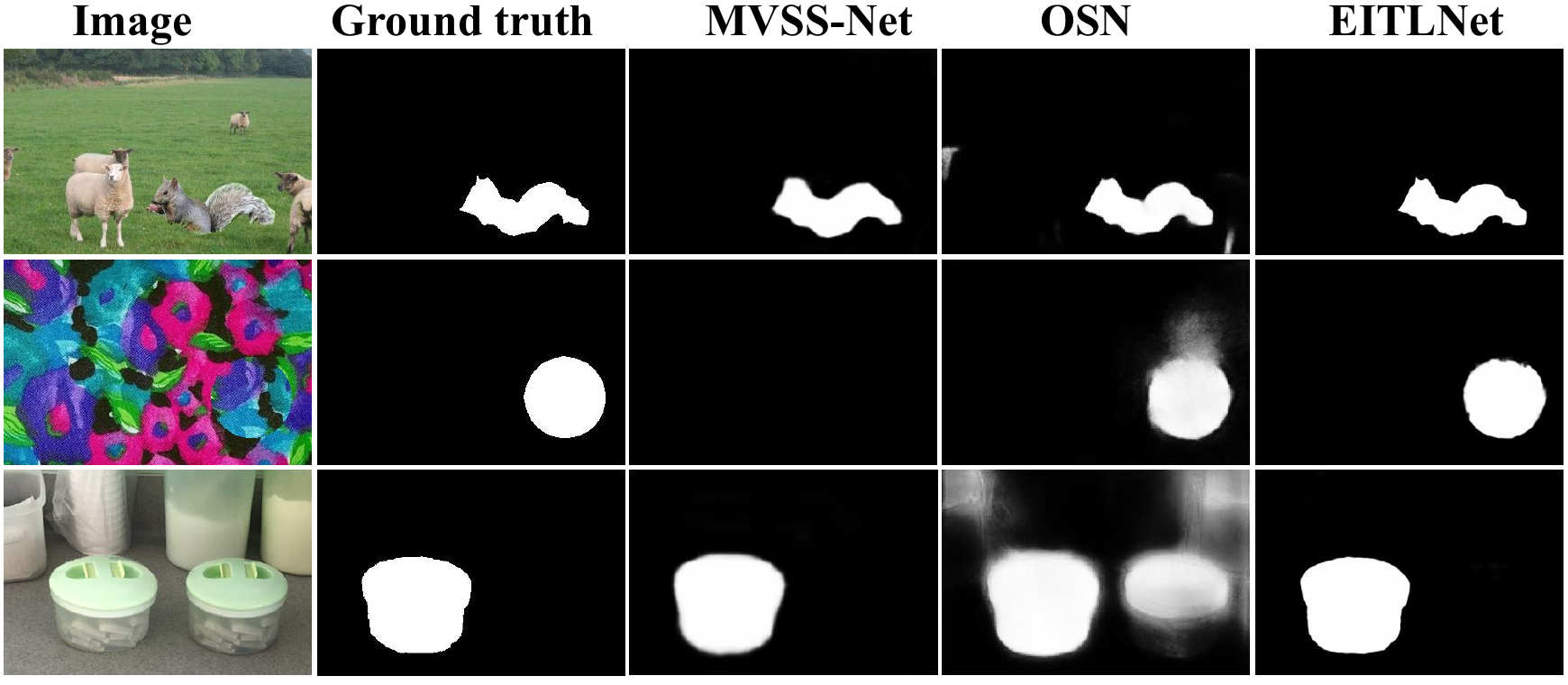}
    \caption{Examples of localization results. From left to right: tampered images, ground truth, localization maps of the localizers including MVSS-Net, OSN and EITLNet.}
    \label{Fig.2.}
\end{figure}
\vspace{-0.5cm}
\begin{table*}[!t]
\caption{Performance comparison of different tampering localization algorithms on different datasets (F1 and IoU). The best results of per test sets are highlighted in bold font. “-” indicates not applicable.}
\vspace{0.2cm}
\begin{adjustbox}{width=\textwidth}
\begin{tabular}{lcccccccccccccc}
\toprule
\multicolumn{1}{c}{\multirow{2}{*}{Methods}} & \multicolumn{2}{c}{Columbia} & \multicolumn{2}{c}{CASIA-v1} & \multicolumn{2}{c}{DSO} & \multicolumn{2}{c}{NIST} & \multicolumn{2}{c}{Coverage} & \multicolumn{2}{c}{IMD} & \multicolumn{2}{c}{Average} \\ 
\cmidrule(r){2-3}  \cmidrule(r){4-5} \cmidrule(r){6-7} \cmidrule(r){8-9} \cmidrule(r){10-11} \cmidrule(r){12-13} \cmidrule(r){14-15}
\multicolumn{1}{c}{}                         & F1            & IoU          & F1            & IoU          & F1         & IoU        & F1          & IoU        & F1            & IoU          & F1         & IoU        & F1           & IoU          \\ \midrule
Mantra-Net, CVPR19\cite{wu2019mantra}                    & 0.357         & 0.258        & 0.130         & 0.086        & 0.332      & 0.243      & 0.088       & 0.054      & 0.275         & 0.186        & 0.183      & 0.124      & 0.228        & 0.159        \\
Noiseprint, TIFS20\cite{cozzolino2019noiseprint}                   & 0.364         & 0.262        & -             & -            & 0.339      & 0.253      & 0.122       & 0.081      & 0.147         & 0.087        & 0.179      & 0.120      & 0.230        & 0.161        \\
DFCN, TIFS21\cite{zhuang2021image}                          & 0.419         & 0.280        & 0.181         & 0.119        & 0.320      & 0.217      & 0.082       & 0.055      & 0.263         & 0.157        & 0.233      & 0.161      & 0.250        & 0.165        \\
MVSS-Net, ICCV21\cite{chen2021image}                      & 0.684         & 0.596        & 0.451         & 0.397        & 0.271      & 0.188      & 0.294       & 0.240      & 0.445         & 0.379        & 0.260      & 0.200      & 0.401        & 0.333        \\
OSN, TIFS22\cite{wu2022robust}                          & 0.707         & 0.608        & 0.509         & 0.465        & \textbf{0.436}      & 0.308      & \textbf{0.332}       & 0.255      & 0.260         & 0.176        & 0.491      & 0.392      & 0.456        & 0.367        \\
EITLNet                                     & \textbf{0.881}         & \textbf{0.851}        & \textbf{0.530}         & \textbf{0.492}        & 0.410      & \textbf{0.319}      & 0.308       & \textbf{0.256}      & \textbf{0.448}         & \textbf{0.371}        & \textbf{0.532}      & \textbf{0.455}      & \textbf{0.518}        & \textbf{0.457}\\       
\bottomrule
\end{tabular}
\end{adjustbox}
\end{table*}

\begin{table*}[!t]

\caption{\leftline{Ablation analysis of our proposed scheme (F1 and IoU). The best results of per test sets are highlighted in bold font.}}
\vspace{0.2cm}
\begin{adjustbox}{width=\textwidth}
\begin{tabular}{lcccccccccccccc}
\toprule
\multirow{2}{*}{Methods} & \multicolumn{2}{c}{Columbia} & \multicolumn{2}{c}{CASIA-v1} & \multicolumn{2}{c}{DSO} & \multicolumn{2}{c}{NIST16} & \multicolumn{2}{c}{Coverage} & \multicolumn{2}{c}{IMD} & \multicolumn{2}{c}{Average} \\
\cmidrule(r){2-3}  \cmidrule(r){4-5} \cmidrule(r){6-7} \cmidrule(r){8-9} \cmidrule(r){10-11} \cmidrule(r){12-13} \cmidrule(r){14-15}
\multicolumn{1}{c}{}   
                         & F1            & IoU          & F1            & IoU          & F1         & IoU        & F1           & IoU         & F1            & IoU          & F1         & IoU        & F1           & IoU          \\ \midrule
Baseline                 & 0.828         & 0.782        & 0.533         & 0.498        & 0.368      & 0.284      & 0.273        & 0.225       & 0.351         & 0.276        & 0.530      & 0.455      & 0.480        & 0.420        \\
Baseline+CAF             & 0.861         & 0.827        & \textbf{0.547}         & \textbf{0.508}        & 0.380      & 0.295      & 0.302        & 0.249       & 0.445         & 0.360        & 0.525      & 0.449      & 0.510        & 0.448        \\
Baseline+CAF+FE          & \textbf{0.881}         & \textbf{0.851}        & 0.530         & 0.492        & \textbf{0.410}      & \textbf{0.319}      & \textbf{0.308}        & \textbf{0.256}       & \textbf{0.448}         & \textbf{0.371}        & \textbf{0.532}      & \textbf{0.455}      & \textbf{0.518}        & \textbf{0.457}        \\
\bottomrule
\end{tabular}
\end{adjustbox}
\end{table*}

\begin{table*}[!t]
\caption{Robustness evaluation results against the social media networks (F1). The datasets are uploaded on Facebook(Fb), Weibo(Wb), WeChat(Wc), WhatsApp(Wa).}
\vspace{0.2cm}
\begin{adjustbox}{width=\textwidth}
\begin{tabular}{lccccccccccccccccccccc}
\toprule
\multirow{2}{*}{Methods} & \multicolumn{4}{c}{Columbia}  & \multicolumn{4}{c}{CASIA-v1}  & \multicolumn{4}{c}{DSO}       & \multicolumn{4}{c}{NIST16}    & \multicolumn{4}{c}{Average}   \\
\cmidrule(r){2-5} \cmidrule(r){6-9} \cmidrule(r){10-13} \cmidrule(r){14-17} \cmidrule(r){18-21}
\multicolumn{1}{c}{}   
                         & Fb    & Wb    & Wc    & Wa    & Fb    & Wb    & Wc    & Wa    & Fb    & Wb    & Wc    & Wa    & Fb    & Wb    & Wc    & Wa    & Fb    & Wb    & Wc    & Wa    \\ \midrule
DFCN               & 0.315 & 0.172 & 0.404 & 0.306 & 0.161 & 0.159 & 0.196 & 0.174 & 0.049 & 0.056 & 0.167 & 0.225 & 0.116 & 0.075 & 0.005 & 0.183 & 0.160 & 0.115 & 0.204 & 0.222 \\
MVSS-Net                 & 0.691 & 0.689 & 0.690 & 0.685 & 0.387 & 0.403 & 0.248 & 0.359 & 0.277 & 0.258 & 0.214 & 0.181 & 0.264 & 0.251 & 0.212 & 0.165 & 0.405 & 0.400 & 0.341 & 0.348 \\
OSN                      & 0.714 & 0.724 & 0.727 & 0.727 & 0.462 & \textbf{0.466} & \textbf{0.405} & 0.478 & \textbf{0.447} & \textbf{0.370} & \textbf{0.366} & 0.341 & 0.329 & 0.294 & 0.286 & 0.313 & 0.488 & 0.464 & 0.446 & 0.465 \\
EITLNet                 & \textbf{0.803} & \textbf{0.896} & \textbf{0.890} & \textbf{0.911} & \textbf{0.473} & 0.453 & 0.364 & \textbf{0.490} & 0.312 & 0.354 & 0.363 & \textbf{0.396} & \textbf{0.370} & \textbf{0.343} & \textbf{0.327} & \textbf{0.349} & \textbf{0.489} & \textbf{0.511} & \textbf{0.486} & \textbf{0.536}\\
\bottomrule
\end{tabular}
\end{adjustbox}
\end{table*}

\subsection{Comparison with other localization algorithms}
Performance of our proposed tampering localization scheme is compared with other state-of-the-art ones, which include Mantra-Net\cite{wu2019mantra}, DFCN \cite{zhuang2021image}, MVSS-Net \cite{dong2022mvss}, and OSN \cite{wu2022robust}. For fair comparison, the DFCN model is retrained on our training dataset. For other methods, we conducted tests using official publicly available code and models. Following the prior work \cite{kwon2022learning}, some NIST16\cite{guan2019mfc} images which can not be tested by ManTra-Net due to GPU memory limit are cropped to 2048×1440 pixels. Table 1 shows the localization results of different methods in terms of F1 and IoU \cite{wu2022robust}. Our method achieves the most superior average performance, outperforming the second-best method OSN by 6.2$\%$ and 9.0$\%$ in terms of F1 and IoU, respectively. It is observed that our proposed scheme always ranks the top two on each dataset, and consistently outperforms all the competing methods in terms of IoU. Fig. 2 provides a more intuitive insight into the actual quality of the results by visualizing examples.

\subsection{Ablation study}
In order to assess the individual impact of the main design choices of our approach, we conduct several ablative experiments. All of these experiments are carried out on the same datasets for training. The baseline model contains a network with RGB and CW-HPF noise stream and concatenate modules, without FE and CAF modules. Table 2 indicates that adopting CAF leads to an average increase of 3.0 $\%$ in the F1 score compared to simple concatenation, along with a 2.8 $\%$ improvement in IoU score. Such results prove the efficiency of our proposed fusion module. The last row of Table 2 illustrates the effect of the FE module. The performance increases to 51.8 $\%$ in F1 and 45.7 $\%$ in IoU on average, which verifies the advantages of the FE module.

\subsection{Robustness evaluation} 
As various lossy operations on social networks pose significant challenges to the robustness of image tampering localization\cite{wu2022robust}, we evaluate the robustness of our model against the postprocessing to the social media platforms. Specificly, the dataset proposed in OSN are evaluated. The four standard forensic datasets are uploaded on Facebook, Weibo, WeChat, and WhatsApp. As illustrated in Table 3, our method shows a consistent gain over all social platforms on Columbia and NIST16 dataset, and achieves competitive performances against the social media networks with other tested platforms. On average, the F1 score has increased on average by 4.7$\%$, 4$\%$, 7.1$\%$ on Weibo, WeChat, WhatsApp.

\section{CONCLUSION}
In this paper, we proposed a novel image tampering localization scheme that leverages a two-branch enhanced transformer encoder and co-attention feature fusion. The feature enhancement module is utilized to strengthen the network representation and the coordinate attention-based fusion module is adopt to combine the multi-scales features. Extensive experiments show that our proposed method achieves superior performance than state-of-the-art models. The EITLNet is verified to be robust against the different social media platforms. In future work, we aim to explore the extension of the proposed method to incorporate noise-based fusion with other approaches.

\small

\bibliographystyle{IEEEbib}
\bibliography{strings,refs}

\end{document}